\documentclass[letterpaper, 10 pt, journal, twoside]{ieeetran}

\usepackage{microtype}
\usepackage{graphicx}
\usepackage{booktabs}
\usepackage{amsmath,amssymb,amsfonts}
\usepackage{mathtools}
\usepackage[T1]{fontenc}
\usepackage{textcomp}
\usepackage{newtxtext}
\usepackage{newtxmath}
\usepackage{multirow}
\usepackage{xcolor}
\usepackage{url}
\usepackage[ruled,vlined,linesnumbered]{algorithm2e}
\usepackage[hidelinks]{hyperref}

\definecolor{leapblue}{RGB}{25,82,140}
\definecolor{leapgreen}{RGB}{0,120,70}
\definecolor{leapred}{RGB}{185,28,45}

\newcommand{\Elatent}{C_{\mathrm{latent}}}
\newcommand{\Eterminal}{E_{\mathrm{terminal}}}

\newcommand{\Etotal}{E_{\mathrm{total}}}
\newcommand{\Phiwm}{\Phi_{\mathrm{LeWM}}}
\newcommand{\R}{\mathbb{R}}
\newcommand{\resultcell}[1]{\makebox[2.4em][r]{$#1$}}
\newcommand{\resultdelta}[3]{\resultcell{#1}\rlap{$^{\textcolor{#2}{#3}}$}}

\title{Latent Energy Action Planning with World Models}

\author{Phu Pham and Aniket Bera \\
\thanks{The authors are with the Department of Computer Science, Purdue University, USA, {\tt\small \{phupham,aniketbera\}@purdue.edu}}
}

\begin{document}
\maketitle

\begin{abstract}
\bfseries
Latent world models support efficient model predictive control from high-dimensional observations, yet optimizing a single learned latent objective can favor action sequences whose decoder-predicted terminal descriptor does not match the goal descriptor. We introduce Latent Energy Action Planning (LEAP), which treats the complete action horizon as a differentiable variable and optimizes it through a frozen LeWorldModel (LeWM). LEAP couples terminal latent goal matching with a terminal-window state energy. Low energy requires the predicted terminal latent to agree with the goal latent and the decoder-predicted terminal descriptor to agree with the goal descriptor. A frozen goal-conditioned proposal initializes the search, a quasi-Newton solver refines actions through the autoregressive rollout, and post-optimization projection enforces the admissible action range. Across four control domains using the officially released LeWM checkpoints, the complete LEAP planning system raises mean success from 77.5\% for LeWM planned with the cross-entropy method (LeWM+CEM) to 94.8\% under a matched protocol, a 17.3-percentage-point improvement, while retaining the frozen LeWM representation and predictor.

\end{abstract}


\section{Introduction}

\IEEEPARstart{L}{atent} world models compress visual observations into predictive states, enabling receding-horizon control without reconstructing pixels. PlaNet~\cite{hafner2019planet} and Dreamer~\cite{hafner2019dreamer} established that compact latent dynamics can support online planning and imagined policy learning. More recent systems scale these ideas across diverse domains. DreamerV3~\cite{hafner2025dreamerv3} learns behavior through imagined experience with one configuration, while TD-MPC2~\cite{hansen2024tdmpc2} performs local trajectory optimization in a learned decoder-free latent model. Joint-embedding predictive architectures replace pixel prediction with feature prediction~\cite{lecun2022path}. I-JEPA~\cite{assran2023ijepa} and V-JEPA~\cite{bardes2024vjepa} instantiate this principle for images and video. DINO-WM~\cite{zhou2025dinowm} uses pretrained patch features for offline world modeling and test-time goal-conditioned action optimization. LeWorldModel (LeWM)~\cite{lewm2026} applies joint-embedding prediction to efficient visual control with a small Vision Transformer and autoregressive latent predictor.

Efficiency does not guarantee that the learned planning objective remains reliable under optimization. A planner actively searches for action sequences on which model error is favorable. This search can lower predicted goal cost outside the action and rollout distributions represented by offline data. MOReL~\cite{kidambi2020morel} and MOPO~\cite{yu2020mopo} motivate pessimistic or uncertainty-penalized model-based offline RL, while BEAR~\cite{kumar2019bear} identifies instability caused by backing up actions outside the data distribution. Our concern is narrower and occurs inside the planner itself. This \emph{optimization-loop loophole} is distinct from ordinary prediction error because average predictive accuracy does not establish accuracy on the trajectories selected by optimization.

\begin{figure}[t]
    \centering
    \includegraphics[width=\columnwidth]{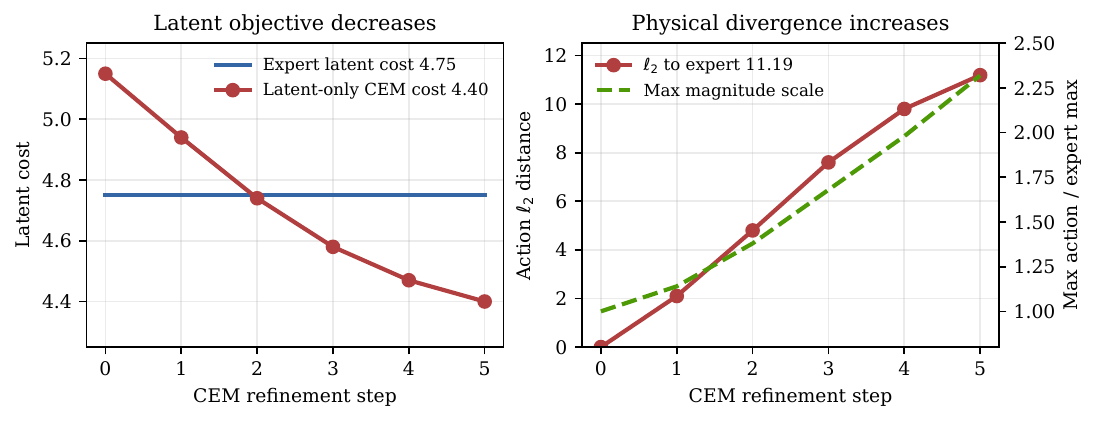}
    \caption{A representative Push-T latent-objective mismatch. Left: latent-only CEM lowers latent cost below the expert reference. Right: the same refinement increases physical action divergence and raises the maximum action magnitude to $2.32\times$ the expert reference.}
    \label{fig:optimization_loophole}
\end{figure}

Figure~\ref{fig:optimization_loophole} shows a representative Push-T latent-objective mismatch during cross-entropy method (CEM)~\cite{rubinstein1999cross} refinement. Latent-only CEM lowers the learned latent cost from the expert reference of 4.75 to 4.40. We define magnitude scale as the candidate maximum absolute action divided by the expert-reference maximum absolute action. Over the same refinement process, the $\ell_2$ distance to the expert controls reaches 11.19 and the magnitude scale reaches $2.32\times$. The opposing trends illustrate a failure mode of latent-only refinement. The improved latent objective coincides with increasing divergence from the demonstrated controls. This example motivates LEAP's optimization-aware planner design.

We address terminal ambiguity at planning time without changing LeWM's representation or predictor. Latent Energy Action Planning (LEAP) treats the complete action horizon as a differentiable variable and backpropagates through the frozen autoregressive rollout. LEAP evaluates the same rollout in LeWM's latent representation and through a decoder-predicted terminal descriptor. The terminal latent goal cost preserves LeWM's native visual objective. The terminal-window state cost requires this low latent cost to be corroborated over the final two predictions. A frozen goal-conditioned proposal centers refinement around a data-fitted action prediction, and post-optimization projection prevents inadmissible controls from being selected. These design choices separate predicted terminal-descriptor disambiguation from action feasibility. The loss equation, decoder architecture, normalization rule, terminal window, and energy weights remain shared across domains.

We evaluate action-gradient planning on Push-T, OGBench-Cube, Reacher, and TwoRooms. All comparisons retain the official LeWM encoder and autoregressive predictor while varying the planner. Paired-start ablations isolate the two energy terms. Under the same protocol, LEAP reaches 94.8\% mean success, compared with 77.5\% for native CEM.

Our contributions are:
\begin{itemize}
    \item We introduce LEAP, a differentiable planner that optimizes complete action horizons through a frozen latent world model using quasi-Newton refinement and post-optimization action projection.
    \item We formulate a terminal energy that couples latent-goal matching with decoder-predicted terminal-state matching, requiring agreement across two complementary representations without retraining the world model.
    \item We conduct an extensive controlled evaluation across four domains and show that LEAP outperforms native LeWM+CEM by 17.3 percentage points while adding only 0.08\,s of planning time per trial.
\end{itemize}

\section{Related Work}

\paragraph{Latent world models.}
Latent dynamics reduce the cost of prediction and planning from images. PlaNet~\cite{hafner2019planet} plans online in a learned latent model, Dreamer~\cite{hafner2019dreamer} differentiates through imagined trajectories, and DreamerV3~\cite{hafner2025dreamerv3} extends imagination-based learning across more than 150 tasks with one configuration. TD-MPC~\cite{hansen2022tdmpc} learns a task-oriented latent model for model predictive control. TD-MPC2~\cite{hansen2024tdmpc2} subsequently scales local latent trajectory optimization across a broader task suite. LeWM~\cite{lewm2026} uses joint-embedding prediction and CEM to obtain efficient visual control. LEAP retains LeWM's learned representation and predictor while redesigning test-time action selection.

\paragraph{Joint-embedding prediction.}
Joint-embedding predictive architectures (JEPAs)~\cite{lecun2022path} predict target features rather than reconstructing pixels. I-JEPA~\cite{assran2023ijepa} applies masked feature prediction to images, and V-JEPA~\cite{bardes2024vjepa} extends the principle to video. DINO-WM~\cite{zhou2025dinowm} predicts pretrained DINOv2 patch features from offline trajectories and optimizes action sequences toward visual goals at test time. LeWM~\cite{lewm2026} instead learns a compact joint-embedding predictor end to end. In both cases the feature space is also the planner's objective. LEAP preserves this latent objective while adding decoder-predicted terminal-descriptor matching during action optimization.

\paragraph{Sampling and gradient planning.}
CEM~\cite{rubinstein1999cross} is widely used because it does not require differentiable dynamics. Sampling-based planning with learned models appears in PETS~\cite{chua2018pets} and visual foresight~\cite{ebert2018visualforesight}. Differentiable MPC~\cite{amos2018differentiablempc} exposes structured controllers to end-to-end gradients through implicit differentiation. Broad search can, however, amplify model error by selecting candidates precisely where the learned objective extrapolates. LEAP instead differentiates a frozen rollout energy directly with respect to the open-loop action tensor and projects the optimized plan into the admissible action range.

\paragraph{Offline support and model exploitation.}
Offline control methods explicitly address decisions that leave the data distribution. MOPO~\cite{yu2020mopo} penalizes uncertain model rollouts, whereas MOReL~\cite{kidambi2020morel} constructs a pessimistic model for policy optimization. In model-free offline RL, BEAR~\cite{kumar2019bear} constrains action selection to reduce bootstrapping error from out-of-distribution actions. LEAP is complementary. It does not fit a policy or value function and does not modify the frozen world model. Instead, it regularizes and projects the action sequence selected by test-time planning.

\paragraph{Energy-based control.}
Energy-based policies and trajectory objectives provide soft mechanisms for combining complementary goal signals. Implicit Behavioral Cloning~\cite{florence2021implicit} models multimodal expert actions with an energy, Composable Energy Policies~\cite{urain2021cep} combine policy factors, and Diffusion Policy~\cite{chi2023diffusionpolicy} performs iterative action refinement through a learned score. LEAP composes latent-goal and terminal-descriptor energies over a multi-step action tensor and differentiates both through a frozen predictive world model.

\section{Problem Formulation}

Let $x_t$ denote the current observation, $g$ a goal observation, and $y_g$ its numerical terminal descriptor. The horizon-$H$ action sequence is $\mathbf a=(a_t,\ldots,a_{t+H-1})\in\R^{H\times d_a}$, where $d_a$ is the action-block dimension. A latent world model encodes $x_t$ into $z_t=\Phiwm(x_t)$ and predicts future latents
\begin{equation}
    \hat{z}_{t+k}=F_{\theta}\!\left(\hat{z}_{t+k-1},e(a_{t+k-1})\right),
    \quad k=1,\dots,H ,
\end{equation}
where $\hat z_t=z_t$, $e$ is the action-block embedding, $F_\theta$ is the frozen autoregressive predictor, and $\Phiwm$ is the frozen observation encoder. Classical latent planning minimizes a learned task cost
\begin{equation}
    \mathbf{a}^{\star}=\arg\min_{\mathbf a}\Elatent(\hat{z}_{t+1:t+H},g).
\end{equation}
For compactness, subsequent costs are written as functions of $\mathbf a$ through the predicted rollout $\hat{\mathbf z}(\mathbf a)=\hat z_{t+1:t+H}$.

\subsection{Motivating Failure of Latent-Only Optimization}

Prediction error describes disagreement on sampled inputs. Latent-only planning creates a distinct risk because the optimizer repeatedly concentrates evaluation on action sequences for which learned error is favorable. Optimization can therefore lower latent cost while moving away from the actions and rollouts represented during training. Figure~\ref{fig:optimization_loophole} shows this mismatch in a representative Push-T trace. The support analysis below formalizes how this failure can arise. It does not assert that every optimized plan leaves the training distribution.

Let $\mathcal{A}_{\mathrm{data}}$ denote the action-sequence support represented in the offline data, and let $\mathcal{Z}_{\mathrm{data}}$ denote the corresponding latent-rollout support. Minimizing only terminal latent error imposes neither
$\mathbf{a}\in\mathcal{A}_{\mathrm{data}}$ nor
$\hat{\mathbf z}(\mathbf a)\in\mathcal{Z}_{\mathrm{data}}$. The planner can therefore select a sequence for which
\begin{align}
    \Elatent(\mathbf{a}^{\star})
    &< \Elatent(\mathbf{a}_{\mathrm{data}}), \\
    \mathbf{a}^{\star}&\notin\mathcal{A}_{\mathrm{data}}
    \quad\text{or}\quad
    \hat{\mathbf z}(\mathbf{a}^{\star})\notin\mathcal{Z}_{\mathrm{data}} .
\end{align}
Here, $\mathbf a_{\mathrm{data}}\in\mathcal A_{\mathrm{data}}$ is a reference sequence.
This condition does not require a particular optimizer. It arises whenever optimization concentrates evaluation on model errors that are rare under the data distribution. Low predicted goal error alone is therefore insufficient evidence that an action plan reaches the intended physical configuration. This motivates complementing the visual latent objective with decoder-predicted terminal-descriptor matching and enforcing admissible controls through post-optimization action projection.

\subsection{Projected Energy-Based Planning Objective}

LEAP augments the terminal latent objective with terminal-state matching. Given $K$ proposal-seeded initializations $\{\mathbf a_0^{(k)}\}_{k=1}^{K}$, it refines and then projects each candidate before selection:
\begin{equation}
\begin{split}
    \tilde{\mathbf a}^{(k)}&=\operatorname{LBFGS}
    \!\left(\Etotal,\mathbf a_0^{(k)}\right),\\
    \mathbf a^{(k)}&=\Pi_{\mathcal A}\!\left(\tilde{\mathbf a}^{(k)}\right),
    \qquad
    k^\star=\arg\min_k \Etotal\!\left(\mathbf a^{(k)}\right).
\end{split}
\end{equation}
Here, $\operatorname{LBFGS}$ denotes the L-BFGS solve, $\mathcal{A}$ is the normalized environment action domain, and $\Pi_{\mathcal A}$ denotes Euclidean projection onto that domain. The selected plan is $\mathbf a^\star=\mathbf a^{(k^\star)}$. Let $D_\psi$ be a frozen decoder from LeWM latent space to a normalized state descriptor. LEAP uses
\begin{equation}
    \Etotal(\mathbf a)=\Elatent(\mathbf a)+\lambda_s\Eterminal(\mathbf a),
\end{equation}
where $\lambda_s>0$ is the terminal-state weight. This energy combines the native visual goal with decoder-predicted terminal-descriptor matching.

Both costs are nonnegative. Thus, for any $\epsilon\geq 0$,
$\Etotal(\mathbf a)\leq\epsilon$ implies
$\Elatent(\mathbf a)\leq\epsilon$ and
$\Eterminal(\mathbf a)\leq\epsilon/\lambda_s$. Because neither term is
negative, a low total cannot mask a large disagreement in either
representation.

The two terms address predicted terminal-descriptor ambiguity through complementary goal representations. The latent goal cost asks whether the terminal prediction resembles the goal in LeWM's representation. The terminal-state cost asks whether the decoder-predicted terminal descriptor matches the goal descriptor under a normalized metric. A low latent score must therefore be corroborated by the decoder prediction instead of serving as the planner's sole evidence of goal attainment. Terminal-state matching does not itself impose membership in $\mathcal{Z}_{\mathrm{data}}$. Neither energy term constrains action magnitude. Post-optimization projection separately enforces admissible controls at the environment interface. The resulting objective combines terminal disambiguation with action feasibility without treating either as a certificate of offline-data support.

\section{Methodology}

\subsection{Differentiable Native-LeWM Rollout}

LEAP changes action selection while leaving LeWorldModel (LeWM) unchanged. The native visual encoder maps the current image and goal image to latent vectors,
\begin{equation}
    z_t=\Phiwm(x_t), \qquad z_g=\Phiwm(g),
\end{equation}
and the frozen autoregressive predictor rolls forward under a candidate action sequence,
\begin{equation}
    \hat z_{t+k}=F_\theta\!\left(\hat z_{t+k-1},e(a_{t+k-1})\right),
    \qquad k=1,\ldots,H.
\end{equation}
The encoder and predictor parameters remain frozen. Unlike native CEM, LEAP retains the computation graph from the predicted latents to the candidate actions. This graph supplies gradients for updating the entire horizon instead of fitting a sampling distribution to ranked candidates.

Native LeWM+CEM treats the frozen rollout model as a black-box evaluator and updates a sampling distribution from ranked action sequences. LEAP instead preserves the rollout computation graph and directly refines the complete action horizon using gradients of the combined terminal energy. Because both planners retain the same LeWM representation and dynamics predictor, their distinction lies in how candidate actions are initialized, refined, and constrained. LEAP therefore converts the pretrained world model from a sampling-based scorer into a differentiable action-planning model.

\begin{figure*}[t]
    \centering
    \includegraphics[width=0.98\textwidth]{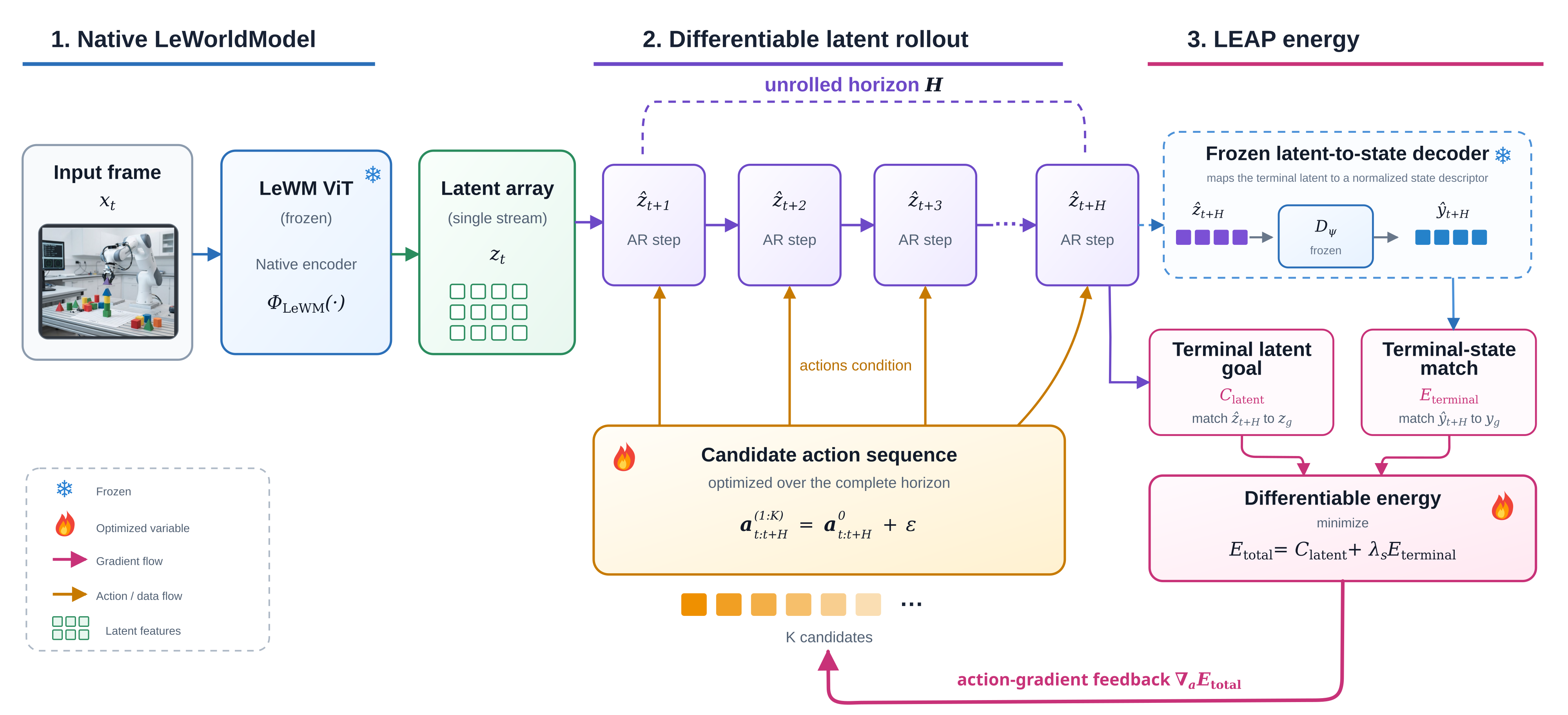}
    \caption{Overview of LEAP. Frozen LeWM encodes the current observation and unrolls each of $K$ candidate action sequences through its autoregressive latent predictor. Candidates are initialized around a goal-conditioned proposal and optimized over the complete horizon, with each action block conditioning its corresponding rollout step. The terminal latent-goal term compares the predicted and goal latents, while a frozen decoder provides normalized descriptors for terminal-state matching. Their sum forms the total energy, whose gradient propagates through the rollout to refine the candidate actions.}
    \label{fig:architecture_overview}
\end{figure*}

Figure~\ref{fig:architecture_overview} summarizes the planner. Both energy terms act on the same frozen rollout, and their sum supplies a single action gradient.

\subsection{Energy-Based Latent Action Planning}

For candidate sequence $\mathbf a\in\R^{H\times d_a}$, the final LEAP design minimizes
\begin{equation}
    \Etotal(\mathbf a)=\Elatent(\mathbf a)+\lambda_s\Eterminal(\mathbf a),
    \qquad \lambda_s=10.
    \label{eq:leap_total}
\end{equation}
The terminal latent goal cost is LeWM's native planning objective,
\begin{equation}
    \Elatent(\mathbf a)=
    \left\|\hat z_{t+H}(\mathbf a)-z_g\right\|_2^2.
    \label{eq:latent_goal}
\end{equation}
It supplies semantic goal direction in the representation on which LeWM was trained.

To predict a complementary terminal descriptor, a frozen decoder $D_\psi$ maps each predicted latent to a standardized state descriptor,
\begin{equation}
    \hat y_{t+k}=D_\psi\!\left(\hat z_{t+k}\right).
\end{equation}
The decoder is fitted from an offline training subset using frozen LeWM embeddings and benchmark state descriptors. Success labels are not used. Each decoder is trained for 50 epochs on 9,000 examples, with 1,000 held-out examples, using the same two-hidden-layer MLP with widths 256--256. Each coordinate is standardized with training-set statistics. The architecture, fitting procedure, coordinate-wise normalization, and loss definition are identical across domains. Only the fitted parameters and dimensionality follow the corresponding frozen checkpoint and descriptor schema.

Let $y_g$ be the standardized goal descriptor and let $L=2$ denote the terminal window. The shared terminal-window state energy is
\begin{equation}
    \Eterminal(\mathbf a)=
    \frac{1}{LD}
    \sum_{k=H-L+1}^{H}
    \left\|D_\psi\!\left(\hat z_{t+k}(\mathbf a)\right)-y_g\right\|_2^2,
    \qquad L=2,
    \label{eq:terminal_state}
\end{equation}
where $D$ is the descriptor dimension. The combined energy turns goal evaluation into cross-representation terminal agreement. A candidate that enters a low latent-cost region but predicts a terminal descriptor inconsistent with $y_g$ remains penalized. Scoring both final predictions further discourages plans that only cross the goal at a single terminal instant. Equation~\eqref{eq:leap_total}, including $L=2$ and $\lambda_s=10$, is unchanged across Push-T, OGBench-Cube, Reacher, and TwoRooms. The algebraic objective contains no domain-specific branches, coordinate masks, or success thresholds.

\begin{figure}[t]
    \centering
    \includegraphics[width=\columnwidth]{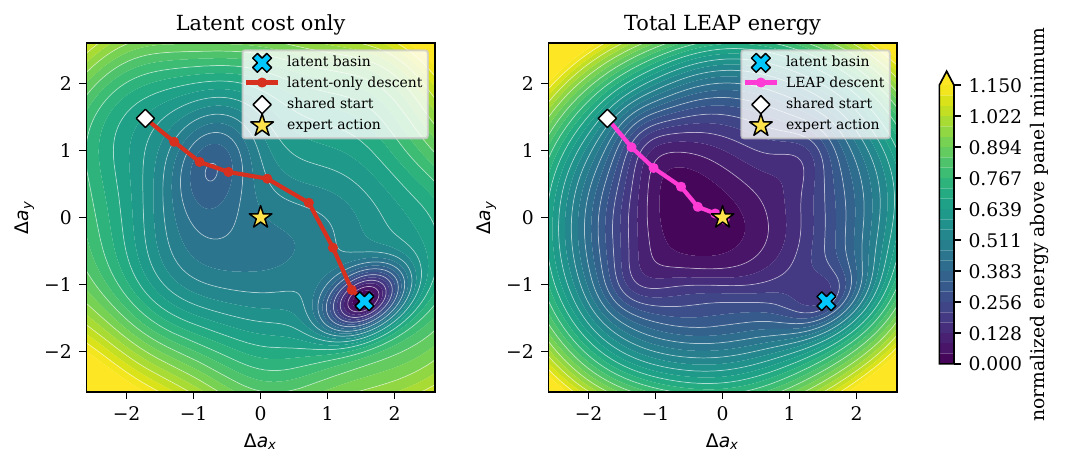}
    \caption{2-D Push-T action-slice energy surface around an expert action. Left: $\Elatent$ contains a misleading off-center basin whose decoder-predicted terminal descriptor is inconsistent with the goal (cyan cross). Right: the total LEAP energy raises this basin and redirects descent toward joint latent and terminal-state agreement (yellow star). Both panels use the same normalized color scale. Irregular local geometry shows that terminal-state matching reshapes rather than simply rescales the action landscape.}
    \label{fig:leap_energy_landscape}
\end{figure}

Fig.~\ref{fig:leap_energy_landscape} provides a geometric analysis of a local 2-D Push-T action slice around an expert action. The asymmetric level sets, secondary basins, and curved descent paths expose the nonconvex geometry faced by action optimization. In the left panel, a misleading latent basin has low cost but is inconsistent with the decoder-predicted terminal descriptor. In the right panel, terminal-state matching raises that basin and redirects descent toward cross-representation agreement. The second term therefore changes the preferred action directions rather than merely rescaling the latent objective.

\subsection{Initialization, Optimization, and Projection}

LEAP initializes candidates with an auxiliary goal-conditioned action proposal trained on offline trajectories and frozen for evaluation. The proposal is a three-hidden-layer MLP with SiLU activations. It maps benchmark-provided planning context to a normalized five-block action sequence. Training minimizes mean-squared error to the corresponding expert sequence using a 95/5 training/validation split, and we retain the minimum-validation-loss checkpoint.

The proposal supplies only the search center and has zero weight in Eq.~\eqref{eq:leap_total}. Its deterministic output defines the first candidate. When $K>1$, seeded perturbations around this center provide additional local starts. This separation keeps the optimized energy compact while avoiding a cold start in the nonconvex action landscape.

For each candidate, LEAP computes
\begin{equation}
    \nabla_{\mathbf a}\Etotal=
    \frac{\partial\Etotal}{\partial \hat{\mathbf z}}
    \frac{\partial \hat{\mathbf z}}{\partial\mathbf a}.
\end{equation}
The final planner uses L-BFGS~\cite{Liu1989OnTL} with learning rate 1.0, ten internal iterations, history size 100, default gradient and parameter-change tolerances of $10^{-7}$ and $10^{-9}$, and strong-Wolfe line search. After the internal solve, post-optimization action projection clamps every candidate to the normalized environment action range. Energy is recomputed on the projected candidates, and selection uses the recomputed values. This enforces admissible controls without changing the internal L-BFGS trajectory.

For $Q$ objective-and-gradient evaluations, $K$ candidate sequences, and rollout horizon $H$, planning requires $\mathcal{O}(QKH)$ predictor steps at fixed model width. L-BFGS line search can invoke more closures than its nominal iteration budget, so we measure synchronized planner time directly rather than treating iteration count as compute. Differentiation retains candidate-specific rollout activations. Memory therefore grows with $K$, $H$, and model width.

After selecting the lowest-energy candidate independently for each environment, LEAP executes five action blocks (25 primitive environment steps) before replanning.

\begin{algorithm}[t]
\caption{LEAP action-gradient optimization}
\label{alg:leap_optimizer}
\DontPrintSemicolon
\KwIn{Observation $x_t$, goal $(g,y_g)$, horizon $H$, starts $K$, L-BFGS budget $M$, weight $\lambda_s$}
\KwOut{Configured receding-horizon prefix}
Encode $z_t\leftarrow\Phiwm(x_t)$ and $z_g\leftarrow\Phiwm(g)$ with native LeWM\;
Initialize $K$ candidates from the frozen action proposal and seeded perturbations\;
Roll out candidates and compute Eqs.~\eqref{eq:latent_goal} and~\eqref{eq:terminal_state}\;
$\tilde{\mathbf a}^{(k)}\leftarrow
\operatorname{L-BFGS}_{M}\!\left(\Etotal,\mathbf a_0^{(k)}\right)$,
$k=1,\ldots,K$\;
$\mathbf a^{(k)}\leftarrow
\Pi_{\mathcal A}\!\left(\tilde{\mathbf a}^{(k)}\right)$\;
$s_k\leftarrow\Etotal\!\left(\mathbf a^{(k)}\right)$\;
Reject the plan if $\{k:s_k<\infty\}=\varnothing$\;
$k^\star\leftarrow\arg\min_{k:s_k<\infty}s_k$\;
\Return $\mathbf a_{t:t+4}^{(k^\star)}$\;
\end{algorithm}

\section{Experimental Evaluation}

\begin{figure*}[t]
    \centering
    \includegraphics[width=0.98\textwidth]{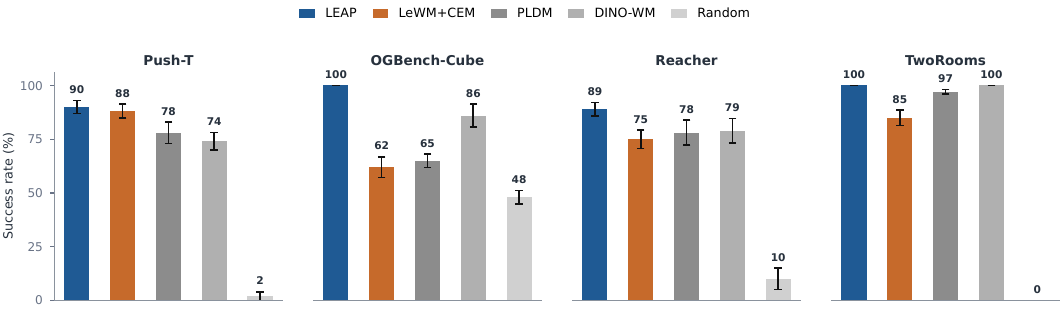}
    \caption{Task success rates (\%). LEAP and LeWM+CEM use our matched controlled 100-trial protocol. Black error bars show one binomial standard error for controlled evaluations and reported variation for contextual baselines.}
    \label{fig:lewm_baselines}
\end{figure*}

\subsection{Domains and Evaluation Protocol}

We evaluate Push-T, OGBench-Cube, Reacher, and TwoRooms using the officially released LeWM checkpoints. The visual encoder and autoregressive predictor remain frozen. Current and goal observations are sampled 25 environment steps apart within the same offline trajectory. Each planner uses a five-transition latent horizon, executes five action blocks (25 primitive environment steps) before replanning, and operates under a 50-step interaction budget. Candidate counts are 32 for Push-T, one for OGBench-Cube, and four for Reacher and TwoRooms. Actions are restricted to $[-2,2]$, except in TwoRooms, which uses $[-1.6,1.6]$.

We select the shared terminal window $L=2$, terminal-state weight $\lambda_s=10$, and optimizer configuration using the paired 50-trial sweeps. These settings are fixed across domains before the final 100-trial evaluation, which uses a different start schedule. No setting is adjusted per task. We also report the corresponding 50-trial final-system results to assess evaluation stability. The LEAP--CEM comparison uses matched sampled starts, task configurations, interaction budgets, official evaluation code, and checkpoints.

Success is binary. Both planners are timed one environment at a time on the same task-assigned GPU. Planning time includes candidate initialization and action optimization but excludes model loading, environment stepping, and visualization. Budgeted steps per trial average all trials, using the first successful primitive step for successful trials and the full interaction budget for failures.

\subsection{Overall Task Performance}

Figure~\ref{fig:lewm_baselines} compares LEAP with our controlled recomputation of native LeWM+CEM. We reproduce the PLDM, DINO-WM, and Random rates reported by LeWM~\cite{lewm2026} for context because we did not rerun those methods. We omit baselines that LeWM did not evaluate on all four tasks, giving every displayed method the same domain coverage.

LEAP reaches 90.0\% on Push-T, 100.0\% on OGBench-Cube, 89.0\% on Reacher, and 100.0\% on TwoRooms, yielding a 94.8\% mean. Under the same protocol, LeWM+CEM reaches 88.0\%, 62.0\%, 75.0\%, and 85.0\%, respectively, for a 77.5\% mean. LEAP changes the planner and adds the terminal-state readout while retaining the frozen LeWM representation and predictor. This controlled comparison evaluates complete planning systems and does not isolate the effects of all planner components individually.

Table~\ref{tab:evaluation_stability} reports both protocols. The 50-trial evaluation yields a 96.5\% mean for LEAP and 77.0\% for LeWM+CEM. LEAP's unweighted mean changes by 1.8 percentage points at 100 trials, and LeWM+CEM changes by 0.5 point. No task-level rate changes by more than four points, leaving every ranking unchanged.

% Generated by scripts/paper/generate_evaluation_stability_table.py. Do not edit.
\begin{table}[!t]
\centering
\caption{Results under the 50- and 100-trial protocols. Entries are success rate $\pm$ one binomial standard error (\%). Mean is unweighted across tasks. Bold identifies LEAP.}
\label{tab:evaluation_stability}
\scriptsize
\setlength{\tabcolsep}{1.5pt}
\renewcommand{\arraystretch}{0.95}
\begin{tabular*}{\columnwidth}{@{\extracolsep{\fill}}clccccc@{}}
\toprule
Trials & Planner & Push-T & \shortstack{OGBench-\\Cube} & Reacher & TwoRooms & Mean \\
\midrule
\multirow{2}{*}{50} & \textbf{LEAP} & $\mathbf{94.0}{\scriptstyle\,\pm\,3.4}$ & $\mathbf{100.0}{\scriptstyle\,\pm\,0.0}$ & $\mathbf{92.0}{\scriptstyle\,\pm\,3.8}$ & $\mathbf{100.0}{\scriptstyle\,\pm\,0.0}$ & $\mathbf{96.5}{\scriptstyle\,\pm\,1.3}$ \\
 & LeWM+CEM & $92.0{\scriptstyle\,\pm\,3.8}$ & $62.0{\scriptstyle\,\pm\,6.9}$ & $72.0{\scriptstyle\,\pm\,6.3}$ & $82.0{\scriptstyle\,\pm\,5.4}$ & $77.0{\scriptstyle\,\pm\,2.9}$ \\
\addlinespace[2pt]
\multirow{2}{*}{100} & \textbf{LEAP} & $\mathbf{90.0}{\scriptstyle\,\pm\,3.0}$ & $\mathbf{100.0}{\scriptstyle\,\pm\,0.0}$ & $\mathbf{89.0}{\scriptstyle\,\pm\,3.1}$ & $\mathbf{100.0}{\scriptstyle\,\pm\,0.0}$ & $\mathbf{94.8}{\scriptstyle\,\pm\,1.1}$ \\
 & LeWM+CEM & $88.0{\scriptstyle\,\pm\,3.2}$ & $62.0{\scriptstyle\,\pm\,4.9}$ & $75.0{\scriptstyle\,\pm\,4.3}$ & $85.0{\scriptstyle\,\pm\,3.6}$ & $77.5{\scriptstyle\,\pm\,2.0}$ \\
\bottomrule
\end{tabular*}
\end{table}

Table~\ref{tab:planner_efficiency} reports matched planning time and environment interaction. Steps per trial average all trials and assign failures the full interaction budget.

% Generated by scripts/paper/generate_planner_efficiency_table.py. Do not edit.
\begin{table}[t]
\centering
\caption{Reported success is in percent. Planning time and environment steps are per trial. Bold identifies LEAP.}
\label{tab:planner_efficiency}
\scriptsize
\setlength{\tabcolsep}{3.5pt}
\begin{tabular}{@{}lccc@{}}
\toprule
Planner & Success (\%) $\uparrow$ & Planning / trial (s) $\downarrow$ & Steps / trial $\downarrow$ \\
\midrule
\textbf{LEAP} & \textbf{94.8} & \textbf{1.28} & \textbf{21.06} \\
LeWM+CEM & 77.5 & 1.20 & 26.41 \\
\bottomrule
\end{tabular}
\end{table}

LEAP improves mean success by 17.3 percentage points and uses 5.35 fewer budgeted steps per trial. Differentiable L-BFGS refinement adds 0.08 s of planning time per trial (1.28 versus 1.20 s). This modest increase accompanies higher success and fewer interactions.

\subsection{Complementarity of the Two Energies}

We compare the complete objective with both single-term variants while holding the checkpoint, proposal, optimizer, candidate count, projection, and paired starts fixed. The latent-goal-only variant differentiates LeWM's native objective through the same planner, while the terminal-state-only variant removes the visual latent term.

% Generated by scripts/paper/generate_final_energy_table.py. Do not edit.
\begin{table}[t]
\centering
\caption{LEAP energy ablation. Paired 50-trial protocol. Task entries and Mean are success rates (\%). Mean is unweighted. Superscripts show percentage-point changes from Full LEAP. Bold marks the best mean.}
\label{tab:leap_components}
\scriptsize
\setlength{\tabcolsep}{2.2pt}
\begin{tabular}{@{}lccccc@{}}
\toprule
Setting & Push-T & \shortstack{OGBench-\\Cube} & Reacher & TwoRooms & Mean \\
\midrule
\textbf{Full LEAP} & \resultcell{\mathbf{94}} & \resultcell{\mathbf{100}} & \resultcell{\mathbf{92}} & \resultcell{\mathbf{100}} & \resultcell{\mathbf{96.5}} \\
Latent goal only & \resultdelta{90}{leapred}{-4} & \resultdelta{94}{leapred}{-6} & \resultdelta{86}{leapred}{-6} & \resultdelta{94}{leapred}{-6} & \resultdelta{91.0}{leapred}{-5.5} \\
Terminal-state only & \resultdelta{34}{leapred}{-60} & \resultdelta{56}{leapred}{-44} & \resultdelta{22}{leapred}{-70} & \resultdelta{96}{leapred}{-4} & \resultdelta{52.0}{leapred}{-44.5} \\
\bottomrule
\end{tabular}
\end{table}

The decomposition reveals an asymmetric complementarity. Latent goal matching is already strong, reaching 90.0\% on Push-T, 94.0\% on OGBench-Cube, 86.0\% on Reacher, and 94.0\% on TwoRooms for a 91.0\% mean. Terminal-state matching alone averages 52.0\%. It is a complementary signal, not a replacement for visual goal identity. Within this paired ablation matrix, their combination is best on every domain and raises the task mean to 96.5\%. With optimization and initialization unchanged, the comparison isolates the objective: semantic latent matching supplies broad goal direction, while terminal-state matching resolves residual decoded error and prevents latent similarity from acting as the sole selection criterion.

\begin{table}[t]
\centering
\caption{First-decision cross-representation agreement. Ratios compare LEAP with latent-goal-only error. Paired lower denotes improved starts. Mean is unweighted across tasks.}
\label{tab:cross_representation}
\scriptsize
\setlength{\tabcolsep}{1.5pt}
\begin{tabular*}{\columnwidth}{@{\extracolsep{\fill}}lccccc@{}}
\toprule
Metric & Push-T & \shortstack{OGBench-\\Cube} & Reacher & TwoRooms & Mean \\
\midrule
Terminal ratio & 0.940 & 0.917 & 0.921 & 0.954 & 0.933 \\
Paired lower & 89\% & 98\% & 96\% & 73\% & 89\% \\
\bottomrule
\end{tabular*}
\end{table}

Table~\ref{tab:cross_representation} compares the first selected plans on
paired starts. LEAP lowers mean decoder-predicted terminal-state error in every
domain, delivering a 6.7\% task-average reduction over latent-goal-only
planning. It selects a lower-error plan on 89\% of starts. These consistent
improvements show that terminal-state matching reliably corrects plans that
latent-only optimization ranks favorably despite decoded terminal mismatch.
With optimizer and initialization fixed, the comparison isolates this
contribution and demonstrates stronger cross-representation goal agreement.

The selected-plan analysis complements task success by showing across domains that the additional energy systematically changes action selection rather than benefiting only a small subset of trials. Together with the component ablation, this result identifies cross-representation terminal agreement as the source of the improvement over latent-goal-only planning.

\subsection{Terminal Regularization Ablations}

We test whether suppressing decoded motion near the horizon improves latent-goal optimization. For $\hat y_{t+k}=D_\psi(\hat z_{t+k})$, define
\begin{equation}
    E_{\mathrm{damp}}(\mathbf a)=
    \frac{1}{(L-1)D}
    \sum_{k=H-L+2}^{H}
    \left\|\hat y_{t+k}-\hat y_{t+k-1}\right\|_2^2,
\end{equation}
which penalizes terminal-window descriptor motion. This ablation-only term has coefficient $\lambda_v$.

% Generated by scripts/paper/generate_loss_design_table.py. Do not edit.
\begin{table}[t]
\centering
\caption{Damping-energy weight ablation with $\lambda_s=0$. Paired 50-trial protocol. Task entries and Mean are success rates (\%). Mean is unweighted. Superscripts show percentage-point changes from latent goal only. Bold marks the best mean.}
\label{tab:damping_weight}
\scriptsize
\setlength{\tabcolsep}{2.2pt}
\begin{tabular}{@{}lccccc@{}}
\toprule
Setting & Push-T & \shortstack{OGBench-\\Cube} & Reacher & TwoRooms & Mean \\
\midrule
$\lambda_v=0$ & \resultcell{90} & \resultcell{94} & \resultcell{86} & \resultcell{94} & \resultcell{91.0} \\
$\lambda_v=0.1$ & \resultdelta{92}{leapgreen}{+2} & \resultdelta{98}{leapgreen}{+4} & \resultdelta{76}{leapred}{-10} & \resultdelta{98}{leapgreen}{+4} & \resultcell{91.0} \\
$\lambda_v=0.3$ & \resultdelta{94}{leapgreen}{+4} & \resultdelta{98}{leapgreen}{+4} & \resultdelta{82}{leapred}{-4} & \resultdelta{98}{leapgreen}{+4} & \resultdelta{93.0}{leapgreen}{+2} \\
$\lambda_v=1$ & \resultcell{90} & \resultdelta{96}{leapgreen}{+2} & \resultdelta{84}{leapred}{-2} & \resultdelta{98}{leapgreen}{+4} & \resultdelta{92.0}{leapgreen}{+1} \\
\textbf{$\lambda_v=3$} & \resultdelta{\mathbf{92}}{leapgreen}{\mathbf{+2}} & \resultdelta{\mathbf{98}}{leapgreen}{\mathbf{+4}} & \resultdelta{\mathbf{88}}{leapgreen}{\mathbf{+2}} & \resultdelta{\mathbf{100}}{leapgreen}{\mathbf{+6}} & \resultdelta{\mathbf{94.5}}{leapgreen}{\mathbf{+3.5}} \\
$\lambda_v=10$ & \resultdelta{92}{leapgreen}{+2} & \resultdelta{96}{leapgreen}{+2} & \resultdelta{78}{leapred}{-8} & \resultdelta{98}{leapgreen}{+4} & \resultcell{91.0} \\
\bottomrule
\end{tabular}
\end{table}

Table~\ref{tab:damping_weight} reports the complete four-domain damping-weight sweep with terminal-state matching disabled. The setting $\lambda_v=3$ is the only tested weight that improves every task over latent goal only and raises the mean from 91.0\% to the best damping-only mean of 94.5\%. Smaller and larger weights regress Reacher, showing that the penalty requires appropriate scaling. We use $\lambda_v=3$ in the composition study of Table~\ref{tab:terminal_composition}.

% Generated by scripts/paper/generate_loss_design_table.py. Do not edit.
\begin{table}[t]
\centering
\caption{Terminal-state weight ablation with $\lambda_v=0$. Paired 50-trial protocol. Task entries and Mean are success rates (\%). Mean is unweighted. Superscripts show percentage-point changes from latent goal only. Bold marks the best mean.}
\label{tab:terminal_state_weight}
\scriptsize
\setlength{\tabcolsep}{2.2pt}
\begin{tabular}{@{}lccccc@{}}
\toprule
Setting & Push-T & \shortstack{OGBench-\\Cube} & Reacher & TwoRooms & Mean \\
\midrule
$\lambda_s=0$ & \resultcell{90} & \resultcell{94} & \resultcell{86} & \resultcell{94} & \resultcell{91.0} \\
$\lambda_s=0.1$ & \resultdelta{92}{leapgreen}{+2} & \resultdelta{98}{leapgreen}{+4} & \resultcell{86} & \resultdelta{98}{leapgreen}{+4} & \resultdelta{93.5}{leapgreen}{+2.5} \\
$\lambda_s=0.3$ & \resultdelta{92}{leapgreen}{+2} & \resultdelta{98}{leapgreen}{+4} & \resultdelta{84}{leapred}{-2} & \resultdelta{98}{leapgreen}{+4} & \resultdelta{93.0}{leapgreen}{+2} \\
$\lambda_s=1$ & \resultdelta{92}{leapgreen}{+2} & \resultdelta{96}{leapgreen}{+2} & \resultdelta{68}{leapred}{-18} & \resultdelta{98}{leapgreen}{+4} & \resultdelta{88.5}{leapred}{-2.5} \\
$\lambda_s=3$ & \resultcell{90} & \resultdelta{98}{leapgreen}{+4} & \resultdelta{68}{leapred}{-18} & \resultdelta{100}{leapgreen}{+6} & \resultdelta{89.0}{leapred}{-2} \\
\textbf{$\lambda_s=10$} & \resultdelta{\mathbf{94}}{leapgreen}{\mathbf{+4}} & \resultdelta{\mathbf{100}}{leapgreen}{\mathbf{+6}} & \resultdelta{\mathbf{92}}{leapgreen}{\mathbf{+6}} & \resultdelta{\mathbf{100}}{leapgreen}{\mathbf{+6}} & \resultdelta{\mathbf{96.5}}{leapgreen}{\mathbf{+5.5}} \\
\bottomrule
\end{tabular}
\end{table}

Table~\ref{tab:terminal_state_weight} disables damping energy and varies $\lambda_s\in\{0,0.1,0.3,1,3,10\}$. Intermediate weights of 1 and 3 reduce mean performance because Reacher drops to 68\%, whereas $\lambda_s=10$ reaches 96.5\% and improves every task over latent goal only. Reacher is particularly sensitive because success requires precise terminal end-effector placement: at intermediate weights, comparably influential latent and terminal-state gradients can steer refinement toward different local basins, whereas $\lambda_s=10$ makes terminal grounding decisive enough to select actions consistent with the decoded goal geometry. This non-monotonic optimization geometry motivates the shared final setting.

% Generated by scripts/paper/generate_loss_design_table.py. Do not edit.
\begin{table}[t]
\centering
\caption{Terminal energy composition. Paired 50-trial protocol. Task entries and Mean are success rates (\%). Mean is unweighted. Superscripts show percentage-point changes from latent goal only. Bold marks the best mean.}
\label{tab:terminal_composition}
\scriptsize
\setlength{\tabcolsep}{2.2pt}
\begin{tabular}{@{}lccccc@{}}
\toprule
Setting & Push-T & \shortstack{OGBench-\\Cube} & Reacher & TwoRooms & Mean \\
\midrule
Latent goal & \resultcell{90} & \resultcell{94} & \resultcell{86} & \resultcell{94} & \resultcell{91.0} \\
Latent + damping & \resultdelta{92}{leapgreen}{+2} & \resultdelta{98}{leapgreen}{+4} & \resultdelta{88}{leapgreen}{+2} & \resultdelta{100}{leapgreen}{+6} & \resultdelta{94.5}{leapgreen}{+3.5} \\
\textbf{Latent + terminal state} & \resultdelta{\mathbf{94}}{leapgreen}{\mathbf{+4}} & \resultdelta{\mathbf{100}}{leapgreen}{\mathbf{+6}} & \resultdelta{\mathbf{92}}{leapgreen}{\mathbf{+6}} & \resultdelta{\mathbf{100}}{leapgreen}{\mathbf{+6}} & \resultdelta{\mathbf{96.5}}{leapgreen}{\mathbf{+5.5}} \\
All three terms & \resultdelta{94}{leapgreen}{+4} & \resultdelta{96}{leapgreen}{+2} & \resultcell{86} & \resultdelta{100}{leapgreen}{+6} & \resultdelta{94.0}{leapgreen}{+3} \\
\bottomrule
\end{tabular}
\end{table}

Finally, Table~\ref{tab:terminal_composition} tests whether the two best auxiliary settings compose. Damping energy alone improves the mean to 94.5\%, but adding $\lambda_v=3$ to the terminal-state term at $\lambda_s=10$ reduces OGBench-Cube from 100\% to 96\%, Reacher from 92\% to 86\%, and the mean from 96.5\% to 94.0\%. The measured regression shows that the two auxiliary terms are not additive under this optimizer. Terminal-state matching penalizes both terminal predictions against the goal, while damping penalizes their separation regardless of goal error. The latter introduces a stationary-state preference that competes with goal matching when the decoder-predicted terminal descriptor is inaccurate. We therefore retain latent goal plus terminal-state matching and exclude damping energy from final LEAP.

\subsection{Optimizer Efficiency}

Table~\ref{tab:leap_optimizer} compares optimizer choices under the final LEAP energy. Task columns and Mean are success rates. Mean superscripts give percentage-point changes from the L-BFGS/10 reference. The L-BFGS/10 row reports the final planner configuration. The remaining rows report the optimizer sweep.

% Generated by scripts/paper/generate_final_optimizer_efficiency_table.py. Do not edit.
\begin{table}[t]
\centering
\caption{Optimizer ablation. Paired 50-trial protocol. Task entries and Mean are success rates (\%). Mean is unweighted. Superscripts show percentage-point changes from L-BFGS/10. Bold marks the best mean.}
\label{tab:leap_optimizer}
\scriptsize
\setlength{\tabcolsep}{2.2pt}
\begin{tabular}{@{}lccccc@{}}
\toprule
Setting & Push-T & \shortstack{OGBench-\\Cube} & Reacher & TwoRooms & Mean \\
\midrule
\textbf{L-BFGS / 10 (ours)} & \resultcell{\mathbf{94}} & \resultcell{\mathbf{100}} & \resultcell{\mathbf{92}} & \resultcell{\mathbf{100}} & \resultcell{\mathbf{96.5}} \\
Adam / 20 & \resultcell{92} & \resultcell{100} & \resultcell{80} & \resultcell{100} & \resultdelta{93.0}{leapred}{-3.5} \\
Adam / 40 & \resultcell{94} & \resultcell{100} & \resultcell{76} & \resultcell{100} & \resultdelta{92.5}{leapred}{-4} \\
L-BFGS / 5 & \resultcell{92} & \resultcell{98} & \resultcell{88} & \resultcell{98} & \resultdelta{94.0}{leapred}{-2.5} \\
L-BFGS / 20 & \resultcell{92} & \resultcell{96} & \resultcell{78} & \resultcell{98} & \resultdelta{91.0}{leapred}{-5.5} \\
\bottomrule
\end{tabular}
\end{table}

Within the paired optimizer study, L-BFGS/10 gives the highest mean success at 96.5\%. The other tested configurations range from 91.0\% to 94.0\%, so L-BFGS/10 remains the final design.

\begin{figure*}[t]
    \centering
    \includegraphics[width=0.98\textwidth]{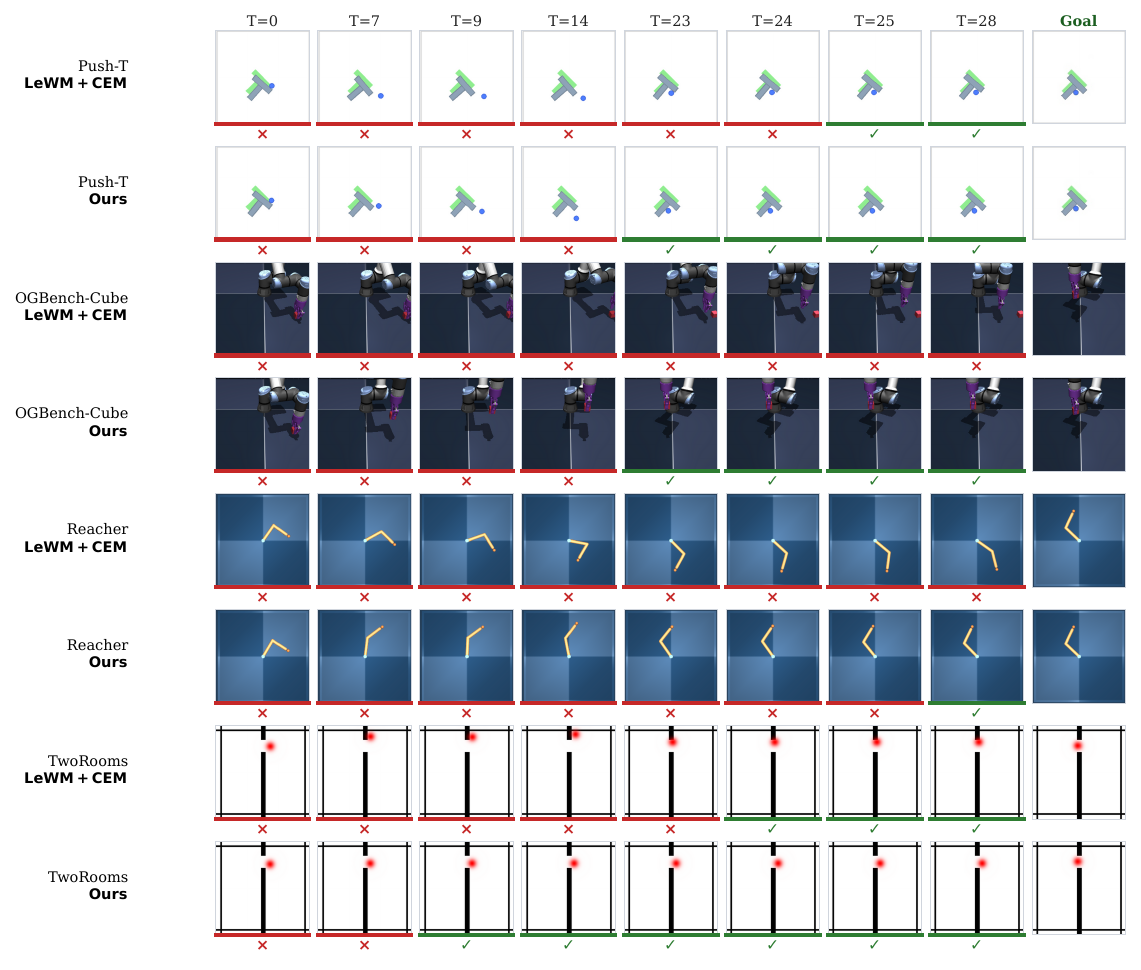}
    \caption{Paired qualitative rollout comparison across four domains. Red crosses denote states before goal attainment, green check marks denote states at or after goal attainment, and the final column shows the goal.}
    \label{fig:rollout_trajectories}
\end{figure*}

\subsection{Qualitative Rollout Analysis}

Fig.~\ref{fig:rollout_trajectories} analyzes paired rollouts for all four domains. The upper row for each task is LeWM+CEM and the lower row is LEAP under the same displayed target. Here, $T$ denotes the primitive environment step. LEAP reaches the goal by $T=23$ on OGBench-Cube and by $T=28$ on Reacher, while LeWM+CEM stays away from the target throughout the displayed sequence. Push-T and TwoRooms instead show paired starts where both planners succeed and LEAP finishes sooner. LEAP requires 23 versus 25 steps on Push-T and 9 versus 24 steps on TwoRooms. For display times after a rollout terminates, the terminal frame is repeated. The status marks make the first successful displayed state explicit. Aggregate success rates across the evaluation schedule appear in Fig.~\ref{fig:lewm_baselines}.

\section{Discussion and Conclusion}

We present LEAP, a differentiable latent-energy planner for action refinement through a frozen world model. Its energy combines latent-goal agreement with decoder-predicted terminal-state matching, while post-optimization projection enforces admissible controls. This formulation defines a differentiable energy over action sequences evaluated through pretrained world-model rollouts without retraining the representation or dynamics model.

Across four manipulation and navigation domains, LEAP raises mean success from 77.5\% with native CEM to 94.8\% while adding 0.08\,s of planning time per trial. Because the official LeWM encoder and predictor remain frozen, the improvement reflects the planning-system design rather than changes to the learned representation or dynamics model. This complete-system comparison does not isolate every planner component.

\textbf{Limitations and Future Directions.} LEAP inherits the frozen world model's limitations on long-horizon and out-of-distribution rollouts. LEAP also uses a decoder trained to predict state descriptors from frozen LeWM embeddings and currently assumes a numerical goal descriptor $y_g$, which may not be available in uninstrumented domains. A promising extension is to replace $y_g$ with a geometry-aware representation $G(g)$ inferred directly from goal images, such as object-centric keypoints or dense visual correspondences learned from unlabeled video. Future work could evaluate this visually grounded formulation under occlusion and domain shift, alongside broader physical and long-horizon evaluation.

\bibliographystyle{IEEEtran}
\bibliography{references}

\end{document}